# Causals: An Open Knowledge Graph for Research on Causal Reasoning

Miaozhe Lin*, Qiongzheng Lin, Zhongwei Liang, Zijun Liu

Causals Lab

## Abstract

Applying causal knowledge to a new event requires connecting the event to relevant entities and the relationships through which its consequences may unfold. We present Causals, an open knowledge graph for research on this form of reasoning. Its initial release, constructed from news reports, contains 369,184 nodes and 1,272,418 edges linking reported events, organizations, commodities, financial instruments, and other entities. Structural relationships, observations, and attributed causal claims retain distinct semantics and links to source records. We describe the construction and representation of the resource and characterize its connectivity and multi-hop retrieval properties. Causals provides shared relational knowledge for studying how language models retrieve relevant context and apply mechanisms to particular events. The dataset is available under CC BY 4.0 at https://causals.org/, and we invite contributions to its ontology, records, and evidence.

## 1 Introduction

Reasoning about the consequences of an event often requires information scattered across many reports. A disruption to a shipping route, for example, can affect industrial inputs, downstream producers, and the markets in which they operate. Tracing these connections requires knowledge of both the event and the relationships among the objects involved. Assessing the consequences further depends on conditions such as inventory, substitutes, timing, and competing influences.

Large language models encode substantial knowledge about causal relationships. Whether they can reliably apply that knowledge beyond familiar examples remains an open question. Causal Parrots argues that fluent causal statements may reflect associations learned from text [1]. Kiciman et al. show that variable names and domain context can help models propose causal relationships [2], whereas Corr2Cause and CLadder examine reasoning under explicit structural or probabilistic constraints [3,4]. Together, these studies distinguish knowledge of familiar causal statements from reasoning with a specified set of relationships.

External knowledge offers a way to supply relationships relevant to a question without updating model parameters. CauseNet collects causal claims expressed on the web [5]. GraphRAG organizes document collections using entity graphs and community summaries [6], while CausalRAG retrieves information through causal paths [7]. These approaches create demand for reusable graph resources on which knowledge selection, representation, and subsequent reasoning can be studied.

News connects general domain knowledge to particular events, organizations, locations, and markets. Reports about different events may refer to the same companies, inputs, or infrastructure, allowing their relationships to accumulate across documents. Constructing such a resource requires collection, extraction, entity reconciliation, and relationship normalization. Sharing the resulting graph allows researchers to study downstream methods on a common dataset rather than repeat this infrastructure work.

* Corresponding author. Email: data@causals.org

We introduce Causals, an open knowledge graph whose initial news-derived release contains 369,184 nodes and 1,272,418 edges. Causals links reported events to concrete entities and records structural relationships, observations, and attributed causal claims. This representation supports the study of event consequences in context: which objects are connected to an event, which relationships may transmit an effect, and what additional information is needed to assess that effect.

Our contribution is a publicly available graph, documentation of its representation and construction, and a reproducible characterization of the released snapshot. We examine referential integrity, source coverage, temporal fields, and connectivity, then measure neighborhood growth under a fixed multi-hop retrieval procedure. These analyses establish the dataset's composition and retrieval properties. The distinction between recorded relationships and causal mechanisms guides both the analysis and the research applications discussed below.

# 2 Related resources and research context

## 2.1 Causal knowledge and formal reasoning

Research on causal reasoning in language models spans several tasks. Causal Parrots [1] questions whether reproducing causal statements demonstrates causal understanding. Kiciman et al. [2] use pretrained domain knowledge to propose relationships between described variables. Corr2Cause [3] tests structural inference from correlational and conditional-independence statements, reducing reliance on familiar variable meanings. CLadder [4] provides causal models and probabilistic information with formally derived answers. Causals complements these task-oriented resources by supplying external relational knowledge about reported events and named entities.

## 2.2 Concept-level claims and contextual graphs

CauseNet [5] contains more than eleven million extracted causal claims between textual concepts, together with extraction provenance. Causals emphasizes connections among particular entities and reported events, including contextual relationships such as supply and ownership alongside causal claims. The resources therefore differ in their unit of representation as well as their size. Table 1 compares their scope with that of two causal-reasoning benchmarks.

*Table 1. Complementary roles of selected resources. Sizes refer to the cited releases or paper, not current live services.*

| Resource | Primary object and size | Research role |
|---|---|---|
| CauseNet [5] | 11.6 million concept-level causal relations | Web-expressed causal knowledge and claim retrieval |
| Corr2Cause [3] | More than 200,000 synthetic text examples | Structural inference from correlation statements |
| CLadder v1.5 [4] | 10,112 formally grounded questions | Association, intervention, and counterfactual reasoning |
| Causals (this release) | 369,184 nodes; 1,272,418 mixed edges | News-derived contextual structure and causal-claim analysis |

## 2.3 Graph-assisted language-model applications

GraphRAG [6] and CausalRAG [7] use graph structure to organize and retrieve information for language models. Causals provides entity names, relation codes, and structural neighborhoods that can serve as inputs to related methods. Its released fields support graph retrieval directly; methods that require source passages need an additional text collection. Section 7 discusses how to use the graph in studies that separate the contribution of retrieved knowledge from that of its representation.

## 3 Resource scope and representation

### 3.1 Unit of analysis

We characterize Causals v0.1.0-2026-10-09. This public snapshot contains 369,184 nodes and 1,272,418 edges after publication filtering from an underlying graph of 452,688 nodes and 1,729,393 edges. All resource statistics refer to this fixed snapshot or to an explicitly defined subset of it.

### 3.2 Nodes and identifiers

Nodes have an id, a set of labels, and an exporter-assigned kind. Labels can overlap: an Event node can also carry Entity. The kind value entity therefore includes events as well as named entities. Source-news records have kind news and label Episodic. Selected non-person entity classes retain names; time and symbol fields are nullable. IDs identify records within this release.

### 3.3 Relationships and semantic distinctions

Edges contain id, source_id, target_id, type, relation, source_episode_ids, and nullable temporal fields. The three edge types are MENTIONS, RELATES_TO, and VARIANT_OF. For RELATES_TO edges, the relation field specifies the semantic relationship. The release contains 39 such codes, or 41 relation values including the other two edge types. Appendix A lists the observed values and their counts.

MENTIONS links a news record to an object, and VARIANT_OF links entity variants. SUPPLIES, PRODUCES, INPUT_TO, and SHIPS_VIA describe structural relationships relevant to event analysis. SHIFTED records an observation. TRIGGERED and STATED_CAUSE represent extracted or attributed causal claims; OTHER groups unclassified relationships. Edge orientation follows the named relationship. Reasoning about a disruption may require traversing that relationship in either direction.

### 3.4 Mechanisms and research assumptions

A relationship identifies a possible route of influence, but its application depends on the event and surrounding conditions. A supplier disruption may affect a buyer's production only if the relevant input is needed, inventories are insufficient, and substitutes are unavailable. Such conditions must come from task context, additional evidence, or an explicit mechanism model. The released graph records relationships and causal claims; it contains neither effect equations nor a complete set of propagation rules.

### 3.5 Example neighborhood

Figure 1 shows four records from the released graph. A shared commodity node connects a producer, a consumer, a shipping route, and a downstream product. An application can retrieve different parts of this neighborhood to examine production exposure, transport dependence, or input requirements. The records can be retrieved by their endpoint names and relation types in the fixed snapshot.

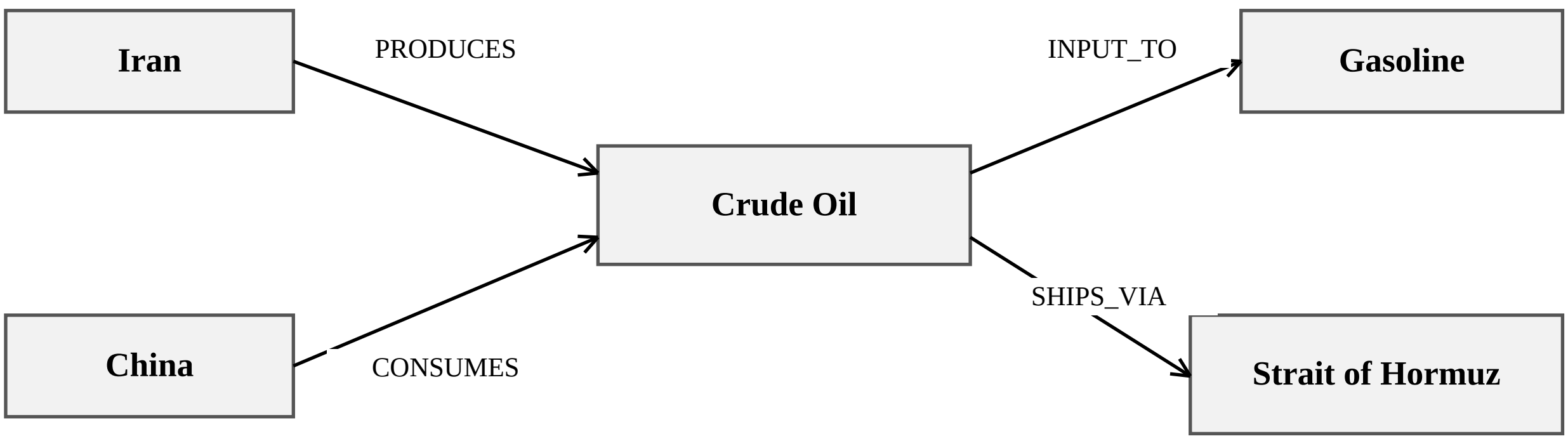

*Figure 1. A neighborhood in Causals v0.1.0-2026-10-09. Labels and arrow directions reproduce the stored relationships.*

# 4 Construction and release

## 4.1 News processing and graph construction

NewsAgent collects reports from configured news sources and extracts article text for processing. The graph-construction pipeline uses language models to identify entities, events, and relationships in that text, with typed structured outputs constrained by the extraction vocabulary. Records are validated before entity reconciliation and graph insertion. Reconciliation combines exact-name lookup with embedding-based candidate retrieval; threshold-based matching handles straightforward cases and model-assisted resolution handles ambiguous matches. Repeated reports can add source references to an existing relationship. Specialized writers supply additional node and relationship types.

Entity reconciliation uses bge-m3 embeddings to retrieve candidate matches. Deterministic links connect news records to mentioned objects and add selected structural relationships. The snapshot combines records accumulated through successive extraction and schema revisions. Source-record links preserve the association between an extracted relationship and its supporting reports.

## 4.2 Publication filtering

The initial release focuses on relational data for graph analysis and retrieval. Person-labelled nodes are excluded because the available annotations do not reliably distinguish public figures from private individuals or identify sensitive personal associations. Uniform exclusion reduces the dissemination of automatically extracted person-level claims. The procedure also excludes non-Person nodes whose names match excluded people and suppresses selected suspicious or overly long names. In total, filtering removed 82,921 Person nodes and 583 name-collision nodes, and suppressed names on 24 further nodes.

Field selection preserves identifiers, selected entity names, relationship codes, source-record links, and temporal fields for structural analysis, while leaving article text and free-text descriptions outside the release. Article headlines, URLs, and event names are also omitted. Source-record IDs support joins within the graph; independent assessment against the original reporting requires the corresponding articles. These choices narrow the available tasks, particularly those involving people or passage-level evidence.

## 4.3 Packaging and reproducibility

Nodes and edges are distributed as gzip-compressed UTF-8 JSON Lines files of 12,224,869 and 60,049,919 bytes, respectively. A manifest supplies file sizes, SHA-256 checksums, relation counts, version information, and the license. Source-record IDs resolve to Episodic nodes in the package. The files can be read without a database installation. Figure 2 summarizes the construction and analysis workflow.

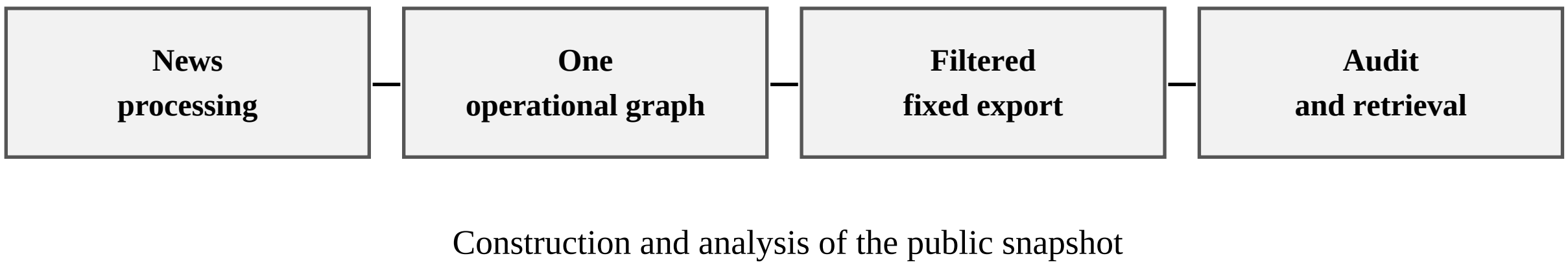

*Figure 2. Construction and analysis workflow. All measurements use the fixed public snapshot.*

# 5 Reproducible audit protocol

## 5.1 Inputs and checks

We analyze the complete node and edge files with audit_snapshot.py. The program recomputes file checksums, tests identifier uniqueness, validates endpoint and source-record references, and compares record counts with the manifest. It also enumerates fields, labels, and relation values. The aggregate results are stored in snapshot_audit.json; file hashes and the full relation inventory appear in Appendix A.

## 5.2 Projections and topology

We use union-find to compute weakly connected components, treating edges as undirected for this calculation. We examine the full graph, RELATES_TO edges, and the subset with relation TRIGGERED or STATED_CAUSE. We refer to the last subset as the claim projection. The full graph uses all released nodes; each projection uses only its incident nodes. Connectivity is computed separately for each view.

## 5.3 Coverage and structural diagnostics

Source coverage is the fraction of records with a nonempty source_episode_ids array. We also test whether every referenced ID resolves to an Episodic node. Temporal coverage is the fraction of records with a non-null, nonempty value in a given field. To assess compatibility with a common reader, we parse each nonempty time string with Python datetime.fromisoformat, replacing a terminal Z with +00:00. We count parsing failures and successfully parsed values without a timezone. This test concerns scalar timestamp formatting, not the factual accuracy of dates.

We count repeated relationships as occurrences after the first of a (source_id, target_id, relation) triple. Records sharing a triple can differ in time or evidence, so we retain them separately. For TRIGGERED and STATED_CAUSE, we test whether both endpoints carry Event labels, following the documented endpoint convention. We also count self-loops. These checks identify structural exceptions for further review.

## 5.4 Deterministic reference retrieval

For the retrieval study, we build an outgoing adjacency index from RELATES_TO edges. Eligible seeds carry Entity but not Event labels and have at least one outgoing RELATES_TO edge. We sort their IDs by SHA-256 digest and select the first 100. From each seed, breadth-first traversal counts distinct other nodes reachable within one, two, and three hops. A visited set excludes the seed and prevents repeated counting through cycles or parallel records.

Traversal uses all RELATES_TO codes without temporal or relevance filtering. The study measures neighborhood expansion for nodes with outgoing semantic edges. We report means, medians, and maxima to capture the influence of highly connected nodes.

# 6 Results

## 6.1 Size and relationship composition

*Table 2. Released records and edge composition. Percentages use all 1,272,418 edges.*

| Measure | Count | Share |
|---|---|---|
| Nodes | 369,184 | -- |
| Edges | 1,272,418 | 100.00% |
| MENTIONS | 703,361 | 55.28% |
| RELATES_TO | 529,414 | 41.61% |
| VARIANT_OF | 39,643 | 3.12% |
| OTHER (within RELATES_TO) | 165,458 | 13.00% |
| TRIGGERED plus STATED_CAUSE | 15,707 | 1.23% |

MENTIONS accounts for 55.28% of edges. The two explicit event-causality codes account for 1.23%, and OTHER contains 165,458 records, or 31.25% of RELATES_TO edges. The node file contains 110,476 source-news records and 258,674 records of kind entity, including 85,935 Event-labelled nodes. A further 34 records have kind other. Names are present on 148,115 nodes. Table 3 gives the nonexclusive label counts.

*Table 3. Nonexclusive domain and event labels. Event can overlap Entity; rows must not be summed as disjoint classes.*

| Label | Nodes |
|---|---|
| Event | 85,935 |
| Company | 52,215 |
| Organization | 31,933 |
| Location | 25,606 |
| Concept | 15,971 |
| Security | 11,563 |
| Software | 8,658 |
| CreativeWork | 6,845 |
| Infrastructure | 3,792 |
| Commodity | 3,179 |
| FinancialIndex | 2,640 |
| EventClass | 946 |
| Currency | 104 |

## 6.2 Integrity, provenance, and claim diagnostics

Both compressed files match the sizes and checksums in the manifest. Recomputed node, edge, relation, and source-coverage counts also agree. We find no duplicate node or edge IDs, dangling endpoints, unresolved source references, references to non-Episodic source nodes, or repeated source IDs within an edge. All fields fall within the export allowlists.

*Table 4. Claim diagnostics. Non-Event endpoints and self-loops are not independent semantic error labels.*

| Relation | Records | With source ID | Not Event to Event | Self-loops |
|---|---|---|---|---|
| TRIGGERED | 8,231 | 7,008 | 650 | 3 |
| STATED_CAUSE | 7,476 | 7,476 | 169 | 0 |
| Combined | 15,707 | 14,484 | 819 | 3 |

Of 529,414 RELATES_TO records, 502,140 (94.85%) have at least one source-record link. Coverage for the claim projection is 92.21%; its 1,223 records without source links all use TRIGGERED. There are 819 claim records whose endpoints do not both carry Event labels. These exceptions may arise from classification history or differences in relation usage and require semantic review. Across all edge types, 8,619 records repeat an earlier source-target-relation triple. Their evidence and temporal fields remain separate.

## 6.3 Temporal completeness and compatibility

*Table 5. Nonempty temporal values, scalar ISO-parser failures, and successfully parsed timezone-naive values. Failures and naive values are separate categories.*

| Field | Nonempty | Coverage | Parser failures | Naive |
|---|---|---|---|---|
| node.valid_at | 195,073 | 52.84% | 853 | 24,099 |
| node.published_at | 85,875 | 23.26% | 0 | 0 |
| edge.as_of | 28,795 | 2.26% | 263 | 3,526 |
| edge.valid_at | 264,439 | 20.78% | 36,812 | 11,059 |
| edge.invalid_at | 39,739 | 3.12% | 8,427 | 5 |

Publication timestamps are present on 85,875 nodes (23.26%) and all pass the scalar parsing test. Node valid_at has higher coverage but includes timezone-naive and unparseable strings. Among 264,439 nonempty edge valid_at values, 36,812 fail parsing. Historical evaluation therefore requires timestamp normalization and a distinction between the time an event occurred and the time its information became available.

## 6.4 Connectivity and neighborhood growth

*Table 6. Weak connectivity of the fixed snapshot and its projections. Nodes are the full universe for the complete graph and incident nodes for the projections.*

| View | Nodes | Components | Largest component |
|---|---|---|---|
| All edges | 369,184 | 18,904 | 337,189 |
| RELATES_TO | 210,848 | 5,149 | 196,214 |
| Claim codes | 18,418 | 5,784 | 4,663 |

The largest weak component contains 91.33% of all released nodes. For RELATES_TO, the largest component contains 93.06% of incident nodes. The claim projection is more fragmented: its largest component contains 4,663 of 18,418 incident nodes (25.32%). Much of the graph's overall connectivity thus comes from relationships outside the explicit causal-claim subset.

*Table 7. Cumulative distinct nodes reached through outgoing RELATES_TO edges from 100 fixed seeds.*

| Maximum hops | Mean | Median | Maximum |
|---|---|---|---|
| 1 | 13.20 | 1.0 | 1,006 |
| 2 | 430.72 | 20.0 | 15,125 |
| 3 | 4,424.65 | 1,255.0 | 37,923 |

Among 76,328 eligible seeds, the selected 100 reach a mean of 13.20 nodes within one hop, 430.72 within two, and 4,424.65 within three. At three hops, the median is 1,255 and the maximum is 37,923. The disparity between the mean and median reflects substantial variation across neighborhoods. Even a short traversal can return more records than a language model can practically use as context, motivating selection by relation, time, or relevance.

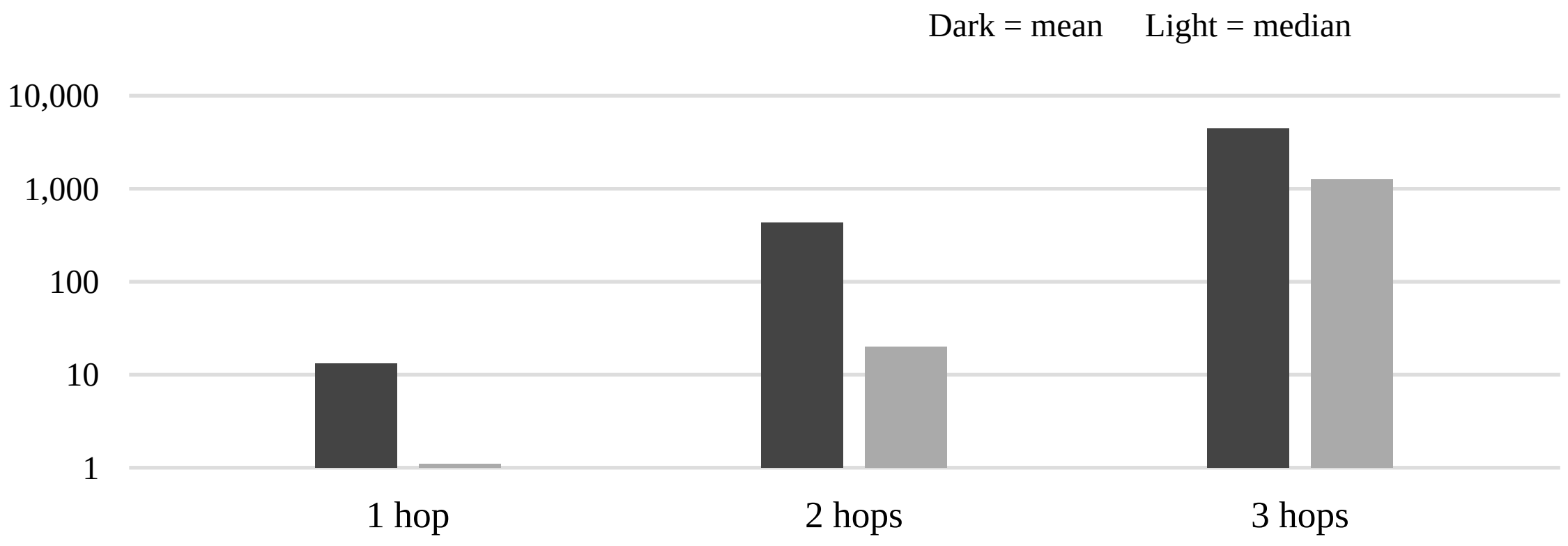

*Figure 3. Mean and median cumulative reachability from the same 100 seeds. The vertical axis is logarithmic.*

### 6.5 Human semantic quality

The released snapshot has no representative estimate of semantic accuracy. The structural checks above identify consistency issues but do not establish whether individual claims are correct. Community review can supply corrections and evidence, and future releases can document quality estimates from a specified sampling and annotation protocol.

## 7 Research applications and reference use

### 7.1 Read-only structural retrieval

A reader can join edges to nodes by ID, filter relation codes, and retrieve a bounded neighborhood. Preserving edge IDs, source-record links, and temporal fields in the result makes each retrieved subset traceable. The procedure in Section 5.4 provides a reproducible starting point using only the downloadable files.

Consider a question about companies exposed to a disrupted input. Retrieval can identify buyers, producers, and related infrastructure around the input. A reasoning method can then assess which connections are relevant under the stated conditions. This separates candidate selection from the assessment of actual consequences and allows each stage to be evaluated independently.

### 7.2 Frozen language models and contextual knowledge

Retrieved records can be supplied as context to a language model without updating its parameters. Selected relationships can also be translated into a task-specific formal model. Useful comparisons include closed-book answering, text retrieval, graph retrieval, and the same retrieved records verbalized as text. Holding evidence content and context budgets constant helps isolate the contribution of graph organization from the benefit of additional knowledge.

### 7.3 Independent downstream datasets

Text-only tasks with independently established reference answers can evaluate entity selection, impact direction, and mechanism explanation. The graph serves as an optional input to the answering system. Task difficulty should be defined independently of access to Causals. In particular, graph distance alone is insufficient to measure causal reasoning depth because mention and classification edges need not represent steps in a causal mechanism.

For financial tasks, future returns provide observable prediction targets. Attributing those returns to particular news requires additional assumptions about contemporaneous influences. Evaluation should specify the information cutoff, asset universe, return horizon, sampling rules, and treatment of abstention.

### 7.4 Application case: market-direction analysis

An operational application uses the graph to connect news-derived event signals to tracked financial assets. Events initialize signed changes in state dimensions such as supply, cost, risk, and earnings expectations. A rule-based engine propagates these changes over typed relationships, subject to endpoint constraints, mechanism conditions, and cycle controls. Asset-level readouts yield an upward or downward direction when sufficient unambiguous information is available. This separates relational knowledge in the graph from the propagation rules used by the application.

We examine a retrospective application to 21 tracked assets over 129 trading dates from 1 May to 3 November 2026. The method retrieves eight-hop neighborhoods and limits each seed to 120 ledger entries per propagation step. Supply and production channels apply an attenuation shift of 0.5 in the propagation score. Event inputs use a 24-hour window ending at 13:00 UTC on evaluation day D. Predictions follow the earnings-expectation readout: upward and downward states map to the corresponding return directions, while missing or conflicting states produce no directional output.

The outcome is the sign of $r1 = close[D]/close[B] - 1$, where B is the most recent earlier date with a price for the asset. Predictions are joined to outcomes by asset and date. A zero return is labelled flat and counts as incorrect for either directional prediction. Let S be the asset-days with both a directional output and an observed outcome. Selective accuracy is the number of correct predictions divided by $|S|$. We also report the scored fraction of the nominal asset-day grid, $|S|/(21T)$, where T is the number of trading dates in the window. This denominator includes every scheduled asset-date cell. The always-up baseline uses the same scored set S.

*Table 8. Exploratory retrospective direction analysis. Grid share is the scored fraction of all 21 assets across the scheduled trading dates. The August-September window is contained in the full window.*

| Window (2026) | Days | Scored / grid | Grid share | Accuracy | Always up | Difference |
|---|---|---|---|---|---|---|
| May 1-Nov 3 | 129 | 872 / 2,709 | 32.2% | 58.26% | 50.00% | +8.26 pp |
| Aug 3-Sep 30 | 42 | 318 / 882 | 36.1% | 61.01% | 53.46% | +7.55 pp |

Across the full window, directional accuracy is 58.26% on 872 scored asset-days, compared with 50.00% for the always-up baseline (Table 8). These cases represent 32.2% of the nominal 2,709-cell asset-day grid. In the nested August-September window, accuracy is 61.01% on 318 scored cases, with a 53.46% baseline. The application illustrates how event signals, typed relationships, and propagation rules can produce measurable asset-level outputs. It uses the operational graph with additional inference rules; the public snapshot supplies the structural resource examined in the preceding sections.

This evaluation is exploratory. Relationships are taken from the operational graph rather than a sequence of historical graph snapshots, and parameters were compared within the same retrospective period. Consequently, the results measure agreement with historical outcomes under this configuration. The nested windows share observations, and returns may be correlated across assets and dates. These results do not isolate the contribution of the graph relative to a language-model-only system or establish prospective trading performance.

### 7.5 Knowledge-assisted discovery

The graph may also supply background knowledge or candidate constraints for causal discovery from numerical data. This application requires mapping entities to measured variables and assessing whether a recorded relationship applies to the population and period under study. Independently validated causal structures are needed to evaluate the resulting discovery procedure.

## 8 Limitations and responsible use

The analysis covers one filtered snapshot. Source selection, newsworthiness, language coverage, and extraction behavior may bias its contents. Incomplete historical collection and model metadata limit assessment of representativeness and reproduction of the original construction process. Identifier stability across later releases has not been established.

The release supports structural analysis more directly than passage-grounded explanation because it lacks article and event text. Typed relationships also leave mechanism conditions and effect magnitudes unspecified. These properties limit which reasoning tasks can be addressed from the public files alone. Label errors, repeated records, and inconsistent timestamp formats may further affect downstream results.

Referential integrity and source coverage measure data consistency and traceability, not semantic correctness. Multiple reports can repeat the same assertion, and a missing edge does not imply that a relationship is absent. The retrieval study measures structural reachability in a fixed 100-seed sample; it has no relevance judgments or causal outcome labels. Its results should therefore be interpreted as workload characterization.

Person filtering reduces the exposure of personal information, but classification errors and sensitive associations may remain. Corrections and removal requests can be sent privately to data@causals.org with the release version and record IDs. The CC BY 4.0 license applies to the rights granted in the dataset; third-party content and applicable privacy rights remain subject to their own terms.

## 9 Availability and community contribution

Causals is available under CC BY 4.0 with attribution to Causals Lab. The dataset, data card, and usage examples are accessible through https://causals.org/. The fixed release is archived under tag graph-v0.1.0-2026-10-09 in the causals-lab/community repository. The accompanying structural analysis takes the released node file, edge file, and manifest as inputs and runs with the Python standard library.

The analysis code and aggregate measurements are available at https://github.com/causals-lab/community/tree/8cabf0854fb2e7a2b25054daeb54615dab63c903/research/causals-paper. The release identifier and file hashes in Appendix A specify the data used to obtain the resource statistics.

Community contributions can address both the ontology and individual graph records. Ontology proposals concern node and relationship types, field definitions, enumerations, and applicability conditions. Record contributions include new subgraphs, corrected nodes and edges, supporting evidence, and counterevidence. Proposals can be submitted through https://github.com/causals-lab/community/issues, following the repository contribution guidelines. Maintainers review proposed changes before incorporating them into a release; sensitive correction requests can be sent privately to data@causals.org.

Planned development includes browser-based contribution and review workflows and additional source types, including textbooks and educational materials. These sources would extend the same graph with general mechanisms and domain relationships alongside reported events. Versioned releases and documented ontology changes will allow users to identify the resource used in a study.

## 10 Conclusion

Causals connects reported events with entities, structural relationships, and attributed causal claims in an openly available news knowledge graph. The first release contains 369,184 nodes and 1,272,418 edges. Our characterization documents its composition, source coverage, temporal properties, and connectivity. Multi-hop traversal reveals rapid and uneven neighborhood growth, highlighting the need to select relevant relationships when constructing model context. The dataset provides a common basis for research on relational retrieval and the use of external knowledge in event reasoning.

## Appendix A Artifact identity and relation inventory

The SHA-256 digests below identify the compressed files used in this study.

```
nodes.jsonl.gz
SHA-256: 96ce2fd336f79a43c6444c73f5d6e6cb2b7ece9e121e6fec2cc52e3781104766

edges.jsonl.gz
SHA-256: 8afd193dbd883d4ae7ec8fbfc3b64137fcd633211b8d56bf894c57a9db08e4aa
```

*Table A1. Complete observed relation inventory, including physical non-semantic relations. Source-linked means a nonempty source_episode_ids array.*

| Relation | Records | Source-linked |
| --- | --- | --- |
| AFFECTS | 6,628 | 6,610 |
| AFFILIATION | 6,286 | 6,285 |
| CLASSIFIED_AS | 811 | 15 |
| COMPETES_WITH | 1,005 | 1,005 |
| CONSTITUTES | 8,536 | 644 |
| CONSUMES | 2,012 | 2,012 |
| CREATED_BY | 6,035 | 6,035 |
| EXPORTS_TO | 1,779 | 1,779 |
| EXPOSED_DOWN | 476 | 462 |
| EXPOSED_UP | 533 | 525 |
| GUARANTEES | 16 | 16 |
| HAS_SECURITY | 2,678 | 367 |
| INDEX_CONSTITUENT | 10,911 | 51 |
| INPUT_TO | 2,016 | 2,016 |
| INSTANCE_OF | 4,408 | 4,261 |
| INVESTS_IN | 7,187 | 7,187 |
| INVOLVES | 221,255 | 221,255 |
| ISSUED_BY | 37 | 0 |
| LOCATED_IN | 6,124 | 5,232 |
| MENTIONS | 703,361 | 0 |
| NOT_TRACKING | 350 | 350 |
| OPPOSES | 5,680 | 5,680 |
| OTHER | 165,458 | 164,245 |
| OWES_TO | 53 | 53 |
| OWNS | 4,533 | 2,830 |
| PART_OF | 4,563 | 4,563 |
| PRODUCES | 6,481 | 6,419 |

| Relation | Records | Source-linked |
|---|---|---|
| REGULATES | 265 | 265 |
| SHIFTED | 18,636 | 18,636 |
| SHIPS_VIA | 570 | 570 |
| SPORTS_MATCH | 393 | 393 |
| STATED_CAUSE | 7,476 | 7,476 |
| STRIKES | 452 | 452 |
| SUPPLIES | 13,713 | 13,713 |
| SUPPORTS | 2,712 | 2,712 |
| TRACKS | 265 | 265 |
| TRANSMITS | 98 | 1 |
| TRIGGERED | 8,231 | 7,008 |
| VARIANT_OF | 39,643 | 0 |
| WORKS_FOR | 677 | 677 |
| WORK_RIGHTS_OF | 75 | 75 |

# Appendix B Field contract and reference traversal

*Table B1. Export field groups. Present keys can contain null values.*

| Record | Fields |
|---|---|
| Node identity | id, labels, kind |
| Node optional values | name, valid_at, published_at, symbol |
| Edge identity and semantics | id, source_id, target_id, type, relation |
| Edge evidence and time | source_episode_ids, valid_at, invalid_at, as_of |

The traversal below uses integer node indices after joining endpoint IDs. The analysis program additionally implements the SHA-256 seed selection described in Section 5.4. Applications that need explanations should retain the corresponding edges and source-record IDs alongside the reachable nodes.

```
visited = {seed}
frontier = {seed}
for depth in range(1, 4):
    next_nodes = {v for u in frontier
                  for v in adjacency.get(u, ())} - visited
    visited.update(next_nodes)
    frontier = next_nodes
    report(depth, len(visited) - 1)
```